\pdfoutput=1

\documentclass[11pt]{article}
\usepackage[review]{ACL2023}

\usepackage{times}
\usepackage{latexsym}
\usepackage{adjustbox}
\usepackage[T1]{fontenc}

\usepackage[utf8]{inputenc}
\usepackage{microtype}

\usepackage{inconsolata}
\usepackage{amsmath}
\usepackage{multirow}

\title{Unveiling Comparative Sentiments in Vietnamese Product Reviews: A Sequential Classification Framework}
\author{
  Ha Le\\
  Fulbright University Vietnam\\
  \\\And
  Bao Tran\\
  Fulbright University Vietnam\\
  \\\And
  Phuong Le\\
  Fulbright University Vietnam\\
  \\\AND
  Tan Nguyen\\
  Fulbright University Vietnam\\
  \\\And
  Dac Nguyen\\
  Athena Studio\\
  \\\And
  Ngoan Pham\\
  VNG Corporation\\
  \\\And
  Dang Huynh\thanks{ \hspace{0.2cm}Corresponding author: \href{mailto:dang.huynh@fulbright.edu.vn}{\nolinkurl{dang.huynh@fulbright.edu.vn}}}\\
  Fulbright University Vietnam\\
}

\date{}

\begin{document}
\nolinenumbers
{\makeatletter\acl@finalcopytrue
  \maketitle
}
\pagestyle{empty}
\begin{abstract}
Comparative opinion mining is a specialized field of sentiment analysis that aims to identify and extract sentiments expressed comparatively. To address this task, we propose an approach that consists of solving three sequential sub-tasks: (i) identifying comparative sentence, i.e., if a sentence has a comparative meaning, (ii) extracting comparative elements, i.e., what are comparison subjects, objects, aspects, predicates, and (iii) classifying comparison types which contribute to a deeper comprehension of user sentiments in Vietnamese product reviews. Our method is ranked fifth at the Vietnamese Language and Speech Processing (VLSP) 2023 challenge on Comparative Opinion Mining (ComOM) from Vietnamese Product Reviews \citep{hoang2023overview}. For reproducing the result, the code can be found at \url{https://github.com/hallie304/VLSP23-Comparative-Opinion-Mining}
\end{abstract}

\section{Introduction}

The landscape of consumer behavior shopping has developed significantly these days, with the surge in online shopping and the omnipresence of e-commerce platforms. Consequently, product reviews have become a ubiquitous source of information, offering a rich tapestry of comparative opinions that manufacturers and consumers find helpful. Understanding these opinions is paramount, serving as a strategic compass for manufacturers to seek insights into their products and for consumers to make wise decisions in the vast amount of choices in the digital marketplace.

Extensive research effort has been spent on comparative opinion mining recently.
However, there has been little effort put in for the Vietnamese language which is a low-resource language compared to English or Chinese. In this paper, we present an approach to address the problem of comparative opinion mining for Vietnamese products. The method consists of three sequential steps which first identify comparative sentences, then extract comparative opinion quadruples, and finally classify comparative features. To cope with natural language processing components, we mainly exploited PhoBERT \citep{phobert}, Electra \citep{clark2020electra}, and Multi-lingual BERT \citep{devlin-etal-2019-bert} to support model ensembling, along with the use of data augmentation techniques to enhance the data quality. These results in an E-T5-MACRO-F1 score of 0.0997, ranked fifth on the private test set.

\begin{figure*}[ht]
\centering
\begin{adjustbox}{width=0.82\textwidth}
\includegraphics{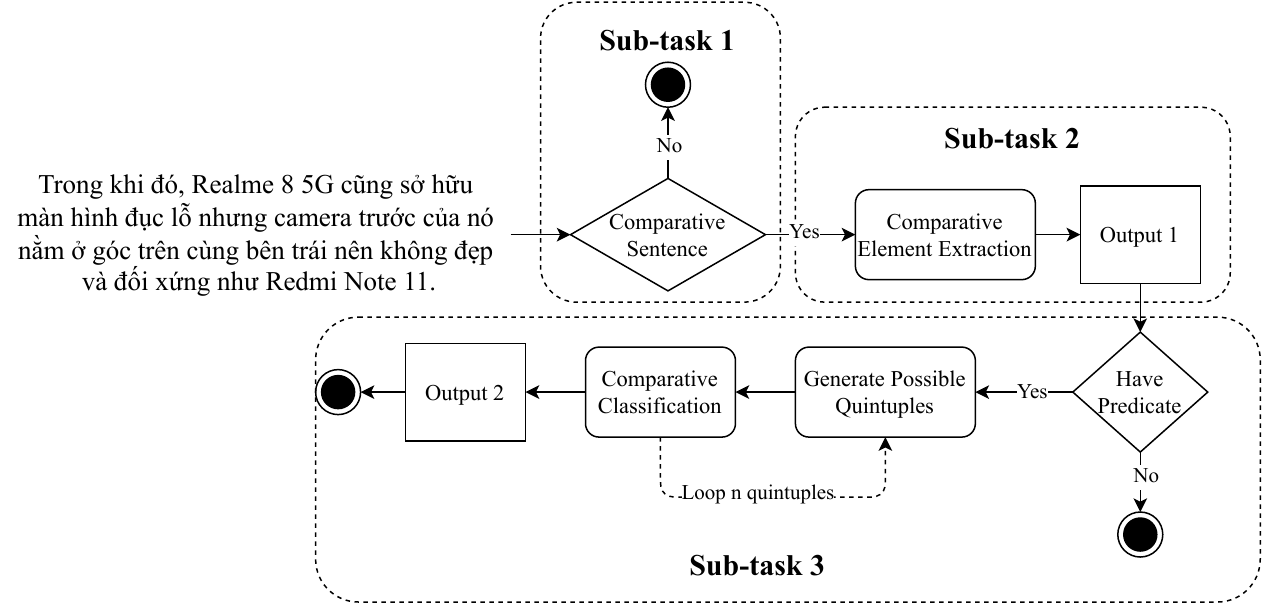}
\end{adjustbox}
\caption{The proposed comparative opinion mining workflow utilizes a Vietnamese product review as input and generates an output quintuple if the review contains comparative sentiments. This quintuple comprises the comparison subject, object, aspect, predicate, and comparison type. There are three distinct sub-tasks: identifying comparative sentences, extracting comparative components from each sentence, and classifying the comparison type of the sentences.}
\label{steps}
\end{figure*}

This paper is structured as follows. Section \ref{sec:relatedwork} provides a comprehensive review of related works, followed by a detailed description of the proposed workflow methodology in Section \ref{sec:methodology}. Section \ref{sec:experiments} presents the experimental results obtained from the application of the methodology, and Section \ref{sec:conclusion} concludes the paper with a summary of key findings and future directions.

\section{Related works}
\label{sec:relatedwork}

A substantial body of research has been devoted to Comparative Opinion Mining (COM) since its inception in 2006. The majority of these studies have focused on English sentences, with proposed solutions primarily addressing three key tasks: Comparative Sentence Identification (CSI), Comparative Entity Extraction (CEE), and Comparative Preference Classification (CPC).

\citet{jindal:2006} pioneered the concept of Comparative Sentence Mining (CSM), introducing a framework for identifying comparative sentences and extracting their comparative quintuples (subject, object, aspect, relation word (or predicate), comparison type). Their initial assumption was that a sentence would only contain one comparative relation, disregarding the possibility of multiple comparative sentiment expressions within a single sentence. Their methods focused on the Comparative Element Extraction (CEE) task, aiming to determine the quintuple and classify its comparison type. They employed a keyword-filtered approach combined with Naive Bayes classification.

In the realm of Comparative Element Extraction (CEE) tasks, several researchers have made significant contributions by introducing novel ideas and techniques, while others have expanded the application of this task to other languages.
Two notable contributions include the Label Sequential Rule (LSR) method proposed by \citet{6201931} for extracting comparative elements and the application of machine learning techniques (SVM) to Chinese sentences by \citet{Wang2015ExploitingML}. For the Comparative Preference Classification (CPC) task, \citet{Jiang2021ConductingPC} introduced their Teardown Joint Sentiment-Topic analysis model. 

Subsequently, \citet{liu:2021} and their colleagues proposed a novel multi-stage neural network approach to address the comparative opinion mining problem. They also introduced an additional task to the COM domain, namely Comparative Opinion Quin-tuple Extraction (COQE). Unlike previous approaches that assumed a single comparative relation per sentence, they formulated sentences as containing multiple quintuples and classified the comparative type based on the element entities within each quintuple. Additionally, \citet{yu2023pretraining} explored the application of pre-trained generative language models (LMs) such as BART or T5 to the comparative opinion mining problem. They trained their models on three distinct tasks: Comparative Answer Generation, Comparative QA Pairs Generation, and Comparative Summary Generation. Both of these works focused on English language data and demonstrated the promising potential of deep neural networks in addressing comparative opinion mining tasks.

\section{Methodology}
\label{sec:methodology}
Our methodology entails constructing a multistage model to address three distinct subtasks: identifying comparative sentences, extracting individual comparative components from each sentence, and labeling the sentences. To enhance the ease of fine-tuning individual sub-tasks, we employ separate models for each sub-task. Specifically, we utilize PhoBERT, Electra, and Multilingual BERT.
\subsection{Preprocessing}
We first conduct a manual review process to identify and rectify any spelling errors or mislabeling inconsistencies. Subsequently, we relabel the dataset and augment it through an upsampling procedure.

Furthermore, the limited size of existing training datasets poses a challenge to model performance. To address this limitation, we incorporate a data augmentation step into our preprocessing pipeline. The specific data augmentation techniques employed are detailed below:
\begin{enumerate}
    \item Isolate and store comparative elements including subjects, objects, aspects, and predicates in separate lists, in which each predicate is matched with a comparative label.
    \item To supplement existing comparative component lists, we incorporate external data sources. Specifically, we extract subjects and objects from curated lists of phone names and brands obtained through web scraping. Additionally, we employ a pre-trained Large Language Model (LLM) to generate novel predicates. ChatGPT was utilized for this task.
    \item Randomly choose a sentence and replace its comparative components with the ones chosen from the above lists, in which the label of the sentence will be determined by the label of the predicate. 
\end{enumerate}

\begin{figure}[ht]
\centering
\begin{adjustbox}{width=0.48\textwidth}
\includegraphics{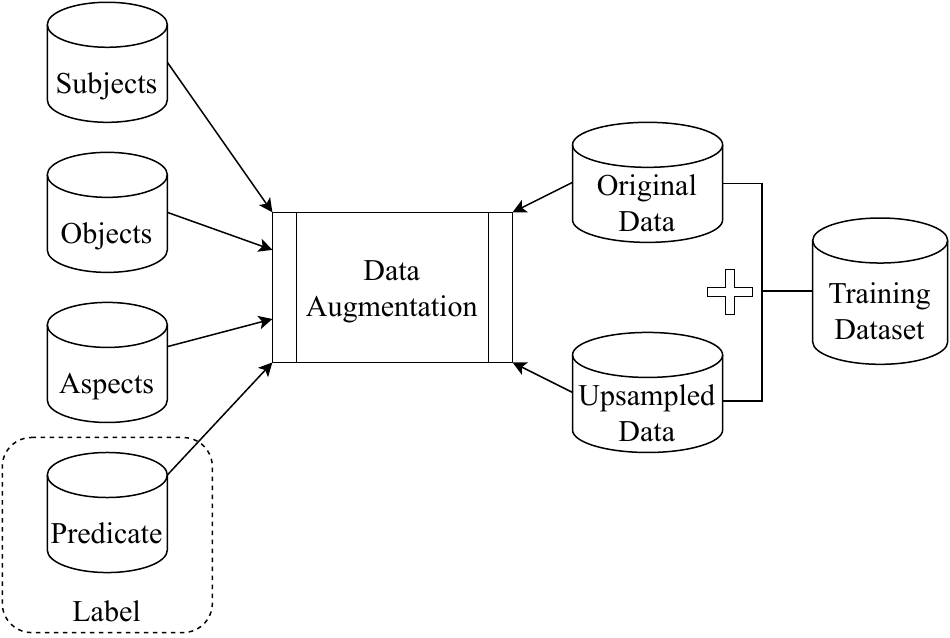}
\end{adjustbox}
\caption{Our training data comprises the combined set of synthetic datasets and the original dataset (after relabeling adjustments). To generate a new dataset, we create dictionaries for subjects, objects, aspects, and predicates. The predicate dictionaries are constructed based on their corresponding comparative types (labels).}
\label{fig:sampling_data}
\end{figure}

\subsection{Feature Engineering}
We combine the generated data with the original data and extract the comparative status, the comparative components, and subsequently, the comparative label of each sentence. The complexity of this task arises from the potential for multiple comparative quintuples (multiple labels) within a single comparative sentence. This necessitates identifying all of these quintuples.
Given a product review sentence containing $n$ words $X = [x_1,\dots,x_n]$, This task aims to determine whether a sentence is comparative and, if so, extract the set of quintuples that represent comparative aspects that contain distinct subjects, objects, aspects, predicates, and types.

\subsection{Modelling}
Our approach employs a multistage model to tackle three distinct sub-tasks: identifying comparative sentences, extracting comparative components from each sentence, and combining comparison elements. Each sub-task is handled by a separate model, facilitating fine-tuning of individual tasks for enhanced performance.
\subsubsection{Sub-task 1: Identifying comparative sentences.}
\begin{figure}[ht]
\centering
\begin{adjustbox}{width=0.5\textwidth}
\includegraphics{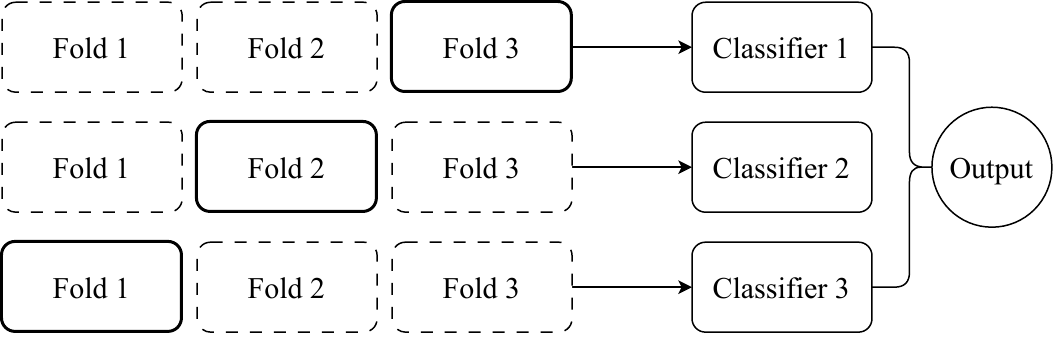}
\end{adjustbox}
\caption{We achieve equal weight for sub-tasks 1 and 3 by combining the outputs of three PhoBERT classifiers. Each classifier is trained on a unique combination of two out of three data folds. For each classifier, the two folds represented with dashed lines are used for training, while the fold with a solid line is used for validation.}
\label{bootstrap}
\end{figure}

For this task, we employed PhoBERT and fine-tuned it with a combination of generated and original data. The sentence content served as the features, while the comparative status was the label. During this phase, we utilized both comparative and non-comparative sentences. Additionally, we implemented 3-fold bootstrapping (see Figure \ref{bootstrap}) to augment the training data for this sub-task. To perform 3-fold bootstrapping, we divided our data into 3 equal folds which are used to train 3 identical PhoBERT classifier. For each classifier, we train it on 2 folds and validate it against the remaining fold. Across the 3 classifiers, each of the 3 folds of the data was used to validate once and was used to train twice. We then combine the 3 trained modifiers by averaging their output logits. This allowed us to maximize the use of our data.

\begin{figure}[h]
\centering
\begin{adjustbox}{width=0.48\textwidth}
\includegraphics{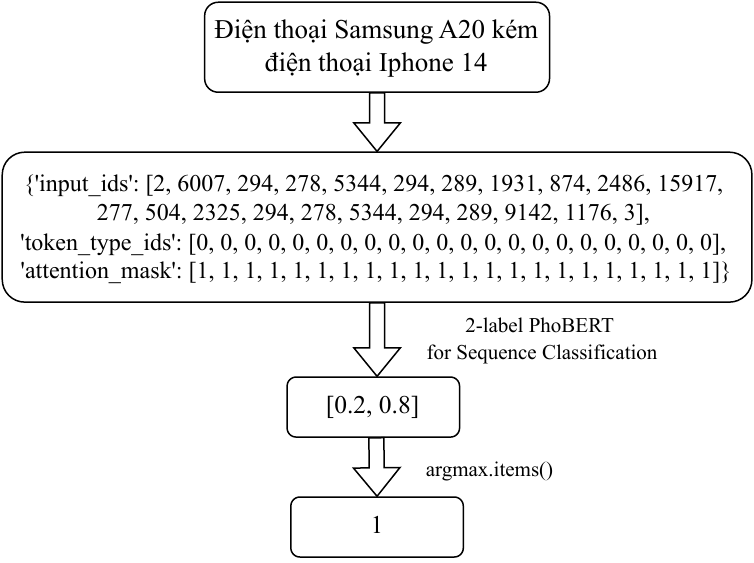}
\end{adjustbox}
\caption{Sub-task 1 - identifying if a sentence is labeled comparative
sentences.}
\label{task1}
\end{figure}

Figure \ref{task1} depicts the sub-task 1 process.

\subsubsection{Sub-task 2: Extracting comparative components of sentences}

This task utilized three models: PhoBERT, Electra, and Multilingual BERT (MBERT). Their outputs were combined using logits-wise ensembling. The features for these models were the sentence content, and the labels were the numerical representation of the comparative components. For this phase, only comparative sentences were used.

Analogous to sub-task 1, the input sentence is first processed by a tokenizer to generate input IDs, token type IDs, and attention masks. Each model employs a distinct tokenizer for this task; however, the output format remains consistent across tokenizers. The encoded input for PhoBERT, Electra, and Multilingual BERT is represented as follows:
$X_{PhoBERT},$ $X_{Electra},$ $ X_{MBERT}$ respectively.

The encoded inputs for each model are subsequently passed through PhoBERT, Electra, and Multilingual BERT models.
The model outputs are represented as $Y_{PhoBERT}, Y_{Electra}, Y_{MBERT}$ respectively.

\begin{figure}[h]
\centering
\begin{adjustbox}{width=0.5\textwidth}
\includegraphics{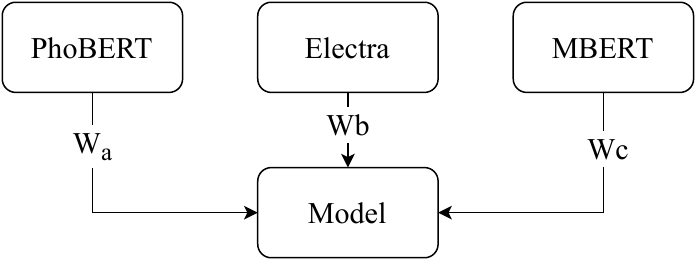}
\end{adjustbox}
\caption{Sub-task 2 - extracting comparative components from sentences.}
\label{ensemble}
\end{figure}

To combine the output tensors of the PhoBERT, Electra, and Multilingual BERT models, we utilize a set of weights determined based on the performance of each model $W_a = 0.2, W_b=0.3,$ and $W_c = 0.5$ respectively (see Figure \ref{ensemble}). These weights were derived through experimentation with different weight combinations for each model output. The final output tensor for sub-task 2 is represented as $Y_{T2}$
\begin{align}
    Y_{T2} = 0.2 \cdot Y_{PhoBERT} + 0.3 \cdot Y_{Electra} \nonumber \\ + 0.5 \cdot Y_{MBERT}
\end{align}

\subsubsection{Sub-task 3: Combining comparison elements}
\begin{figure}[h]
\centering
\begin{adjustbox}{width=0.5\textwidth}
\includegraphics{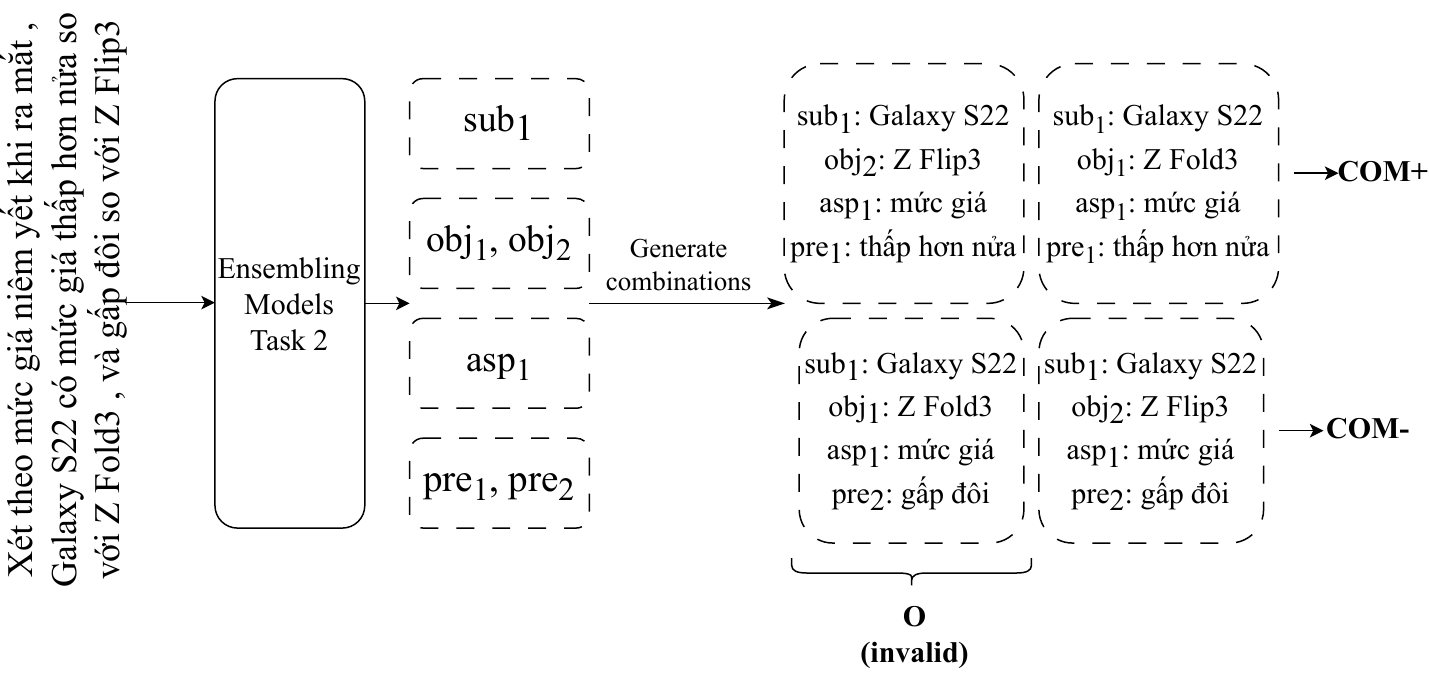}
\end{adjustbox}
\caption{Sub-task 3 - Generating all possible quadruples of the four entities and combining comparison elements.}
\label{task3}
\end{figure}

Sub-task 2 yielded four sets of comparative elements for comparative sentences, denoted as
$S_{sub} = \{sub_1, \cdots, sub_i\},$ $S_{obj} = \{obj_1, \cdots, obj_j\},$ $S_{asp} = \{asp_1, \cdots, asp_k,\}, $ $S_{pre} = \{pre_1, \cdots, pre_l\}$
We then generate all possible quadruple combinations $S_{qua}$ using these four sets. Figure \ref{task3} details an example of the process.
\begin{align}
    S_{qua} = \{(sub_1, obj_1, asp_1, pre_1), \cdots, \nonumber \\
    (sub_i, obj_j, asp_k, pre_l)\}
\end{align}

Similar to sub-task 1, the input is fed into the PhoBERT tokenizer to be encoded into input IDs, token type IDs, and attention masks. Here we use the 9-label sequence classification PhoBERT model along with the bootstrap technique (see Figure \ref{bootstrap}) to obtain the final results. 

\section{Experiments}
\label{sec:experiments}
\subsection{Experimental settings}
There are five experiments that differ from each other in the use of dataset versions and the number of output classes for sub-task 1.
\begin{itemize}
    \item Experiment 1: Dataset version 2 (data summary in Table~\ref{table2}) with bootstrapping and training sub-task 1 with 9 classes (8 comparison components and no comparison label).
    \item Experiment 2: Dataset version 2 without bootstrapping, and training sub-task 1 with 2 classes.
    \item Experiment 3: Dataset version 2 with bootstrapping, and training sub-task 1 with 2 classes.
    \item Experiment 4: Dataset version 3 (data summary in Table~\ref{table3}), without bootstrapping, and training sub-task 1 with 9 classes.
    \item Experiment 5: Dataset version 3 with bootstrapping, and training sub-task 1 with 9 classes.
\end{itemize}

All of the experiments were run on the same computer, which has a Tesla P100 GPU with 16 GB of memory, an Intel Xeon CPU running at 2 GHz, and 32 GB of RAM. All three sub-tasks were trained using the AdamW optimizer with a learning rate of 3e-5, a batch size of 32, and 15 epochs.

\subsection{Dataset cleaning}

The original dataset contains unexpected characters such as non-breaking and zero-width spaces. These special characters should be considered during word indexing, necessitating adjustments to the NER (Named Entity Recognition) process for each sentence during the tokenization phase.

\begin{table}[h]
\begin{tabular}{|c|c|}
\hline
\textbf{Error Type}              & \textbf{Number} \\ \hline
1. Excessively long predicate    & 30              \\ \hline
2. Wrong or missing quintuple    & 25              \\ \hline
3. Redundant or wrong components & 19              \\ \hline
\end{tabular}
\caption{Summary of the three common error types in the original dataset.}
\label{error_summary}
\end{table}

\begin{figure}[h]
\centering
\begin{adjustbox}{width=0.48\textwidth}
\includegraphics{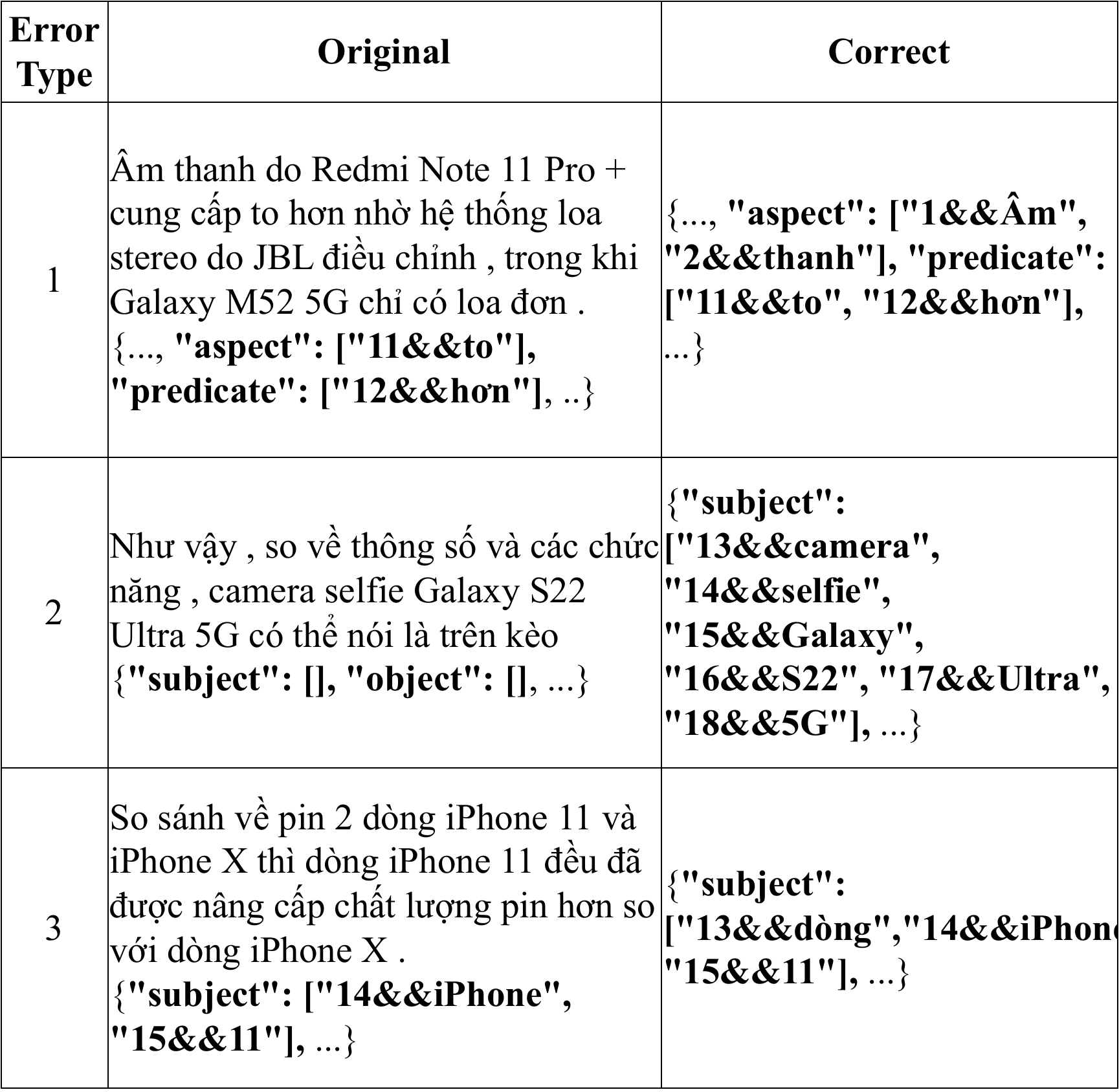}
\end{adjustbox}
\caption{Some examples of relabeling dataset.}
\label{relabeled_examples}
\end{figure}

\begin{table}[h]
    \centering
    \begin{tabular}{p{\dimexpr.07\linewidth-2\tabcolsep-1\arrayrulewidth}|c|c|c}
     &Type &Number &Percent \\
     \hline
     \parbox[t]{4mm}{\multirow{4}{*}{\rotatebox[origin=c]{90}{Sentence}}}& Multi-Comparative & 222 & 27.47\% \\
     &Mono-comparative & 586  & 72.53\% \\
     \cline{2-4}
     &Comparative & 808 & 19.4\% \\
     &Non-comparative & 3359 & 80.6\%  \\
     \hline
     \parbox[t]{4mm}{\multirow{8}{*}{\centering\rotatebox[origin=c]{90}{Label}}}& DIF & 58 & 5.32\% \\
     &EQL & 287  & 26.35\% \\
     &SUP+ & 107 & 9.83\% \\
     &SUP- & 5 & 0.46\%  \\
     &SUP & 4 & 0.37\%  \\
     &COM+ & 500 & 45.91\%  \\
     &COM- & 107 & 9.83\%  \\
     &COM & 21 & 1.93\%  \\
     \hline         
     \parbox[t]{4mm}{\multirow{4}{*}{\rotatebox[origin=c]{90}{Element}}}& Subject entity & 428 &  \\
     &Object entity & 356  &  \\
     &Aspect entity & 565 &  \\
     &Predicate entity& 630 &  \\
\end{tabular}
\caption{Dataset version 1 analysis.}
\label{table1}
\end{table}

The dataset exhibits an imbalance between comparative and non-comparative sentences, as evident in Table \ref{table1}. The number of non-comparative sentences significantly outnumbers the number of comparative ones. Moreover, the prevalence of single-meaning comparative sentences overshadows the instances of multiple comparative meanings. This skewness poses a challenge for the overall model's detection capabilities. If left unaddressed, it could lead to overfitting or underfitting during training. The imbalance is particularly acute for the SUP, SUP-, COM, and DIF classes, where data samples are scarce compared to other classes. Our training process utilizes this dataset, designated as version number one, hence the table's name.

In the second version of the dataset, we generate synthetic data samples while preserving the distribution of the original dataset version.

\begin{table}[h]
    \centering
    \begin{tabular}{p{\dimexpr.07\linewidth-2\tabcolsep-1.3333\arrayrulewidth}|c|c|c}
     &Type &Number &Percent \\
     \hline
     \parbox[t]{4mm}{\multirow{4}{*}{\rotatebox[origin=c]{90}{Sentence}}}& Mono-Comparative & 4070 & 73.8\% \\
     &Multi-comparative & 1445  & 26.2\% \\
     \cline{2-4}
     &Comparative & 5515 & 47.03\% \\
     &Non-comparative & 6212 & 52.97\%  \\
     \hline
     \parbox[t]{4mm}{\multirow{8}{*}{\centering\rotatebox[origin=c]{90}{Label}}}& DIF & 410 & 5.61\% \\
     &EQL & 1788  & 24.47\% \\
     &SUP+ & 334 & 4.57\% \\
     &SUP- & 288 & 3.95\%  \\
     &SUP & 308 & 4.22\%  \\
     &COM+ & 2980 & 40.77\%  \\
     &COM- & 854 & 11.68\%  \\
     &COM & 346 & 4.73\%  \\
     \hline         
     \parbox[t]{4mm}{\multirow{4}{*}{\rotatebox[origin=c]{90}{Element}}}& Subject entity & 6426 &  \\
     &Object entity & 4440  &  \\
     &Aspect entity & 6215 &  \\
     &Predicate entity& 7256 &  \\
    \end{tabular}
    \caption{Dataset version 2 summary}
    \label{table2}
\end{table}
Table \ref{table2} shows the updated version 2 of our dataset. In this version, we approximately multiply each class of comparative sentence by six with the minor classes being approximately 300 each. This helps the model perform better in detecting minor classes. 

Despite advancements, the model's ability to identify minor classes remains a challenge. The public test results revealed that the model struggled with SUP- classes, prompting the development of the third dataset version to address this limitation.

\begin{table}[h!]
    \centering
    \begin{tabular}{p{\dimexpr.07\linewidth-2\tabcolsep-1.3333\arrayrulewidth}|c|c|c}
     &Type &Number &Percent \\
     \hline
     \parbox[t]{4mm}{\multirow{4}{*}{\rotatebox[origin=c]{90}{Sentence}}}& Mono-Comparative & 2740 & 74.07\% \\
     &Multi-comparative & 960 & 25.95\% \\
     \cline{2-4}
     &Comparative & 3700 & 37.33\% \\
     &Non-comparative & 6212 & 62.67\%  \\
     \hline
     \parbox[t]{4mm}{\multirow{4}{*}{\centering\rotatebox[origin=c]{90}{Label}}}& DIF & 536 & 11.07\% \\
     &EQL & 557  & 11.48\% \\
     &SUP+ & 597 & 12.31\% \\
     &SUP- & 610 & 12.58\%  \\
     &SUP & 545 & 11.24\%  \\
     &COM+ & 770 & 15.88\%  \\
     &COM- & 597 & 12.31\%  \\
     &COM & 638 & 13.15\%  \\
     \hline         
     \parbox[t]{4mm}{\multirow{4}{*}{\rotatebox[origin=c]{90}{Element}}}& Subject entity & 4357 &  \\
     &Object entity & 2745  &  \\
     &Aspect entity & 3969 &  \\
     &Predicate entity& 4827 &  \\
    \end{tabular}
    \caption{Dataset version 3 summary}
    \label{table3}
\end{table}

Table \ref{table3} represents the most refined dataset version in our work. This iteration ensures that all classes have an equal number of samples, leading to improved performance in detecting two classes that previously lacked sufficient data.
\subsection{Evaluation result}

The models are ranked by their macro-averaged F1 score over the comparison types, shown in Table \ref{table:f1final}. This metric, E-T5-MACRO-F1, provides a more balanced view of performance than individual precision and recall values. Further details on precision and recall are provided in Tables \ref{table:pfinal} and \ref{table:rfinal} respectively.

\begin{table}[h]
\centering
\begin{tabular}{lc}
\hline
\textbf{Teams} & \textbf{MACRO-F1}   \\
\hline
thindang & 0.2373\\
pthutrang513 & 0.2300\\ 
thanhlt998 & 0.2131\\
duyvu1110 & 0.1119\\
\textbf{Our team} & \textbf{0.0997}\\
kien-vu-uet & 0.0975 \\
\hline
\end{tabular}
\caption{E-T5-MACRO-F1 comparison between our team and the others on the private test set.}
\label{table:f1final}
\end{table}

\begin{table}[h]
\centering
\begin{tabular}{lc}
\hline
\textbf{Teams}  & \textbf{MACRO-P}   \\ 
\hline
thindang & 0.2862\\
thanhlt998 & 0.2093\\
pthutrang513 & 0.2021\\
\textbf{Our team}  & \textbf{0.0968} \\
duyvu1110 & 0.0964 \\
kien-vu-uet & 0.0933 \\ 
\hline
\end{tabular}
\caption{E-T5-MACRO-P comparison between our team and the others on the private test set.}
\label{table:pfinal}
\end{table}

\begin{table}[!h]
\centering
\begin{tabular}{lc}
\hline
\textbf{Teams}  & \textbf{MACRO-R}   \\ 
\hline
pthutrang513 & 0.2718\\
thindang & 0.2216\\
thanhlt998 & 0.2199\\
duyvu1110 & 0.1375 \\
\textbf{Our team}  & \textbf{0.1065} \\
kien-vu-uet & 0.1057 \\ 
\hline
\end{tabular}
\caption{E-T5-MACRO-R comparison between our team and the others on the private test set.}
\label{table:rfinal}
\end{table}
These findings offer valuable insights into boosting model performance. Notably, upsampling the dataset, especially for under-represented classes like DIF and SUP-, demonstrably improves recognition accuracy for these classes. Additionally, bootstrapping proves to be a potent technique, particularly when data resources are limited.
\section{Conclusion}
\label{sec:conclusion}
In this paper, we have introduced a sophisticated sequential classification framework meticulously designed to analyze comparative sentiments embedded within Vietnamese product reviews. This approach effectively captures the dataset's inherent diversity, ensuring the model is thoroughly exposed to a comprehensive range of linguistic patterns and syntactic structures.

For future work, as an attempt to improve the workflow, one can think of extending the models to handle the scenario in sub-task 2 where a word can function as both the subject and object in different quintuples. This can be achieved by modifying the input tensor for Name Entity Recognition (NER) to explicitly encode the roles of words in quintuples, enabling the model to make more informed decisions and achieve superior accuracy in relation extraction tasks.

\bibliographystyle{acl_natbib}
\bibliography{anthology,custom}
\end{document}